\documentclass[11pt]{article}

\usepackage[final]{acl}
\usepackage{amsmath}
\usepackage{enumitem}
\usepackage[skip=2pt]{caption}
\usepackage{titlesec}
\titlespacing*{\section}
  {0pt}   
  {2ex} 
  {0.8ex} 
 
\usepackage{times}
\usepackage{latexsym}
\usepackage{hyperref}
\usepackage{pgfplots}
\pgfplotsset{compat=1.18}
\usepackage{fancyvrb}
\usepackage{float}
\DefineVerbatimEnvironment{PromptBlock}{Verbatim}{
  fontsize=\footnotesize,
  baselinestretch=1,
  breaklines=true,
  breakanywhere=true
}
\DefineVerbatimEnvironment{SchemaBlock}{Verbatim}{
  fontsize=\footnotesize,
  baselinestretch=1,
  breaklines=true,
  breakanywhere=true
}
\usepackage{booktabs}
\usepackage{tabularx}
\usepackage{array}
\usepackage{enumitem}
\newcolumntype{Y}{>{\raggedright\arraybackslash}X}
\usepackage{listings}
\usepackage{xcolor}

\usepackage[T1]{fontenc}

\usepackage[utf8]{inputenc}

\usepackage{microtype}

\usepackage{inconsolata}

\usepackage{graphicx}

\title{Semantic Grading of Written Answers in Low-Resource Language Bangla Using a Fine-Tuned Lightweight Language Model}

\author{First Author \\
  Affiliation / Address line 1 \\
  Affiliation / Address line 2 \\
  Affiliation / Address line 3 \\
  \texttt{email@domain} \\\And
  Second Author \\
  Affiliation / Address line 1 \\
  Affiliation / Address line 2 \\
  Affiliation / Address line 3 \\
  \texttt{email@domain} \\}

\author{
 \textbf{Meherun Farzana}\thanks{Equal contribution},
 \textbf{Aniket Joarder}\footnotemark[1],
\\
 \textbf{Mahmudul Hasan},
 \textbf{Md.\ Mosaddek Khan},
\\
 Computer Science and Engineering, University of Dhaka
 \\
  \texttt{\{meherun-2020815615, aniket-2020115658,}\\
\texttt{mahmudul-2019917803\}@cs.du.ac.bd, mosaddek@du.ac.bd}
}

\begin{document}
\maketitle
\begin{abstract}
Bangla is among the world’s most widely spoken languages, yet it remains underserved in educational NLP research. In many remote and rural regions, access to qualified subject teachers is limited, and written answers are consequently graded largely by hand, restricting timely and consistent feedback. Automatic assessment is challenging because semantically correct responses can vary substantially in surface form. We present a bilingual (Bangla–English) evaluation system designed for low-resource educational settings that prioritizes semantic correctness over lexical overlap.\footnote{Demo Video: \href{https://youtu.be/mDZme6uaQzo}{\texttt{YouTube}}}\footnote{Live Website: \href{https://cognifyq.com}{\texttt{cognifyq.com}}} Our approach fine-tunes a lightweight language model to grade each response using the question, reference answer, and student answer, producing a numeric score and concise, context-grounded feedback suitable for classroom deployment. We also construct a synthetic bilingual dataset to enable controlled training and evaluation. Across proprietary and open-source LLMs evaluated under a unified protocol, our QLoRA-tuned Qwen3-8B confirms consistent improvement by producing the most leakage-resistant feedback (\textsc{RoRa} = 0.819) in synthetic evaluation and the strongest agreement with human scores ($\rho = 0.936$, $\mathrm{MAE} = 0.725$) in a dedicated human study.
\end{abstract}

\section{Introduction}
Written answers better reflect what students understand than multiple-choice items, but grading them at scale is difficult, as responses vary widely in wording and quality. Teachers may assign different scores to borderline cases where it is unclear whether a student deserves full credit, partial credit, or no credit. The need for a reliable evaluation system is especially acute for Bangla, the \emph{5th most spoken native language in the world} \citep{ethnologue-ten-largest-2025}. However, no widely established end-to-end system exists for grading Bangla written answers in high school exam settings, particularly for handwritten scripts. In Bangladesh, large class sizes and exam-driven assessment create substantial grading workloads, and in many remote and rural regions limited access to qualified subject teachers further restricts timely and consistent feedback. A practical grading system for Bangla written answers could reduce turnaround time, improve scoring consistency, and free teachers to focus on instruction and targeted student support.


Prior work on automatic text evaluation began with surface-overlap metrics such as BLEU \citep{papineni2002bleu}, and later shifted toward semantic similarity and learned evaluators, including BERTScore \citep{zhang2020bertscore} and UniEval \citep{zhong2022unieval}. More recently, LLM-based meta-evaluation frameworks such as G-Eval \citep{liu2023geval} and multi-agent deliberation in ChatEval \citep{chan2024chateval} aim to better align with human preferences, but they are primarily designed for open-ended NLG tasks rather than curriculum-grounded answer correctness.

In educational NLP, systems have moved beyond scoring to feedback. SAF \citep{filighera2022saf} pairs short answers with explanatory feedback, IFlyEA \citep{gong2021iflyea} provides essay scoring with review and recommendations, and Chinese narrative comment generation systems \citep{zhang-etal-2022-automatic-comment} generate teacher-like comments for essay segments. Parallel work such as PEEP-Talk \citep{lee2023peep} focuses on language learning through dialogue practice and feedback. Despite these advances, existing approaches exhibit key limitations: (1) they overwhelmingly target resource-rich languages (e.g., English, Chinese, German, Spanish), with limited attention to low-resource languages such as Bangla; (2) they rarely evaluate curriculum-grounded scientific or factual answers where semantic correctness is central; (3) feedback modules are often template-driven, limiting adaptability and faithfulness; and (4) many rely on large proprietary models or complex pipelines that are difficult to deploy in low-resource educational settings.

Building on these gaps, we present an end-to-end bilingual (Bangla and English) written answer evaluation system that prioritizes semantic correctness over surface overlap and remains practical under low-resource constraints. Our core contributions are:

\begin{itemize}[leftmargin=*, itemsep=0pt, topsep=-0.5pt, parsep=0pt, partopsep=0pt]
\item We fine-tune an open-weight grader (Qwen3-8B) using task-specific QLoRA and evaluate each response from a fixed context tuple (\emph{question, reference answer, student answer}), ensuring contextual grounding and reducing reliance on surface-level word matching.
\item We introduce a synthetic bilingual benchmark with four answer variants (\textsc{Good/Partial/Plausible/Bad}) per question, explicitly designed to stress-test semantic correctness and partial-credit assignment.
\item We support both typed and handwritten Bangla answers via an HTR module, GraDeT-HTR \citep{hasan-etal-2025-gradet}, and produce a numeric score with concise, context-grounded feedback.
\item We conduct a dedicated human evaluation study and compare our system against proprietary and open-source LLMs under a unified protocol. Our QLoRA-tuned Qwen3-8B produces the most faithful feedback (\textsc{RoRa} = 0.819) synthetically and achieves the strongest alignment with the human-graded scores ($\rho=0.936$, $\mathrm{MAE}=0.725$), outperforming the base model, while remaining deployable in low-resource classrooms.
\end{itemize}

\section{System Overview and Architecture}
\label{sec:system}

\begin{figure}[t]
  \centering
  \includegraphics[width=0.78\columnwidth]{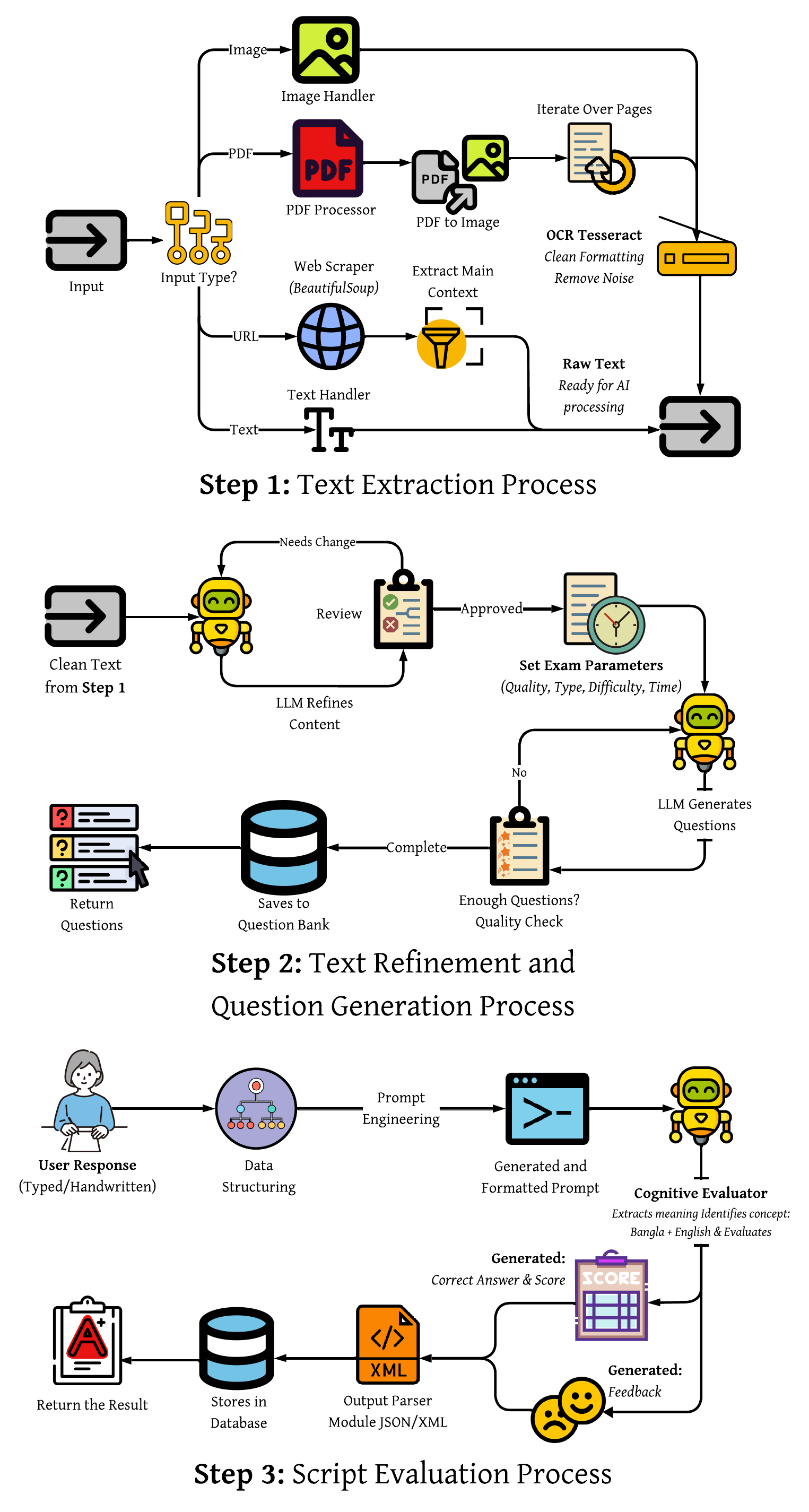}
  \caption{System architecture: (1) text extraction from different types of sources, (2) question generation with refined text, and (3) answer script evaluation producing scores with feedback.}
  \label{fig:cognifyq_method}
\end{figure}

We propose an end-to-end bilingual assessment system for written answers designed for realistic exam inputs and low-resource deployment. The pipeline consists of three stages (Figure~\ref{fig:cognifyq_method}): (i) input-conversion and context extraction, (ii) constraint-controlled question generation, and (iii) answer evaluation with scoring and feedback.


\subsection{Input-conversion and Bangla Handwriting}
The system supports four input formats: PDF, image, URL, and plain text, as well as a book from the Book Management. For PDFs, the converter processes pages sequentially and extracts embedded text; for scanned pages and images, it runs OCR. General-purpose OCR is largely tuned for printed Latin script and often performs poorly on Bangla. Hence, for state-of-the-art performance and low-resource deployment, we use GraDeT-HTR \citep{hasan-etal-2025-gradet}, a Bangla handwriting text recognition (HTR) component, to handle Bangla handwritten responses. For English handwriting, we use Google Vision OCR. For URLs, it retrieves the main article content while filtering boilerplate such as navigation and ads. After conversion into raw text, we apply lightweight normalization (Unicode cleanup and whitespace standardization) to produce a consistent text representation for downstream modules.

\subsection{Constraint-controlled Question Generation}
\label{subsec:method_qgen}
When teachers provide source materials, the system can generate exam-ready questions under explicit constraints (subject/topic, difficulty, time limit of taking the exam). Generated questions are wrapped in a lightweight “quality gate” that checks format validity, coverage, and near-duplicates; failed questions are revised and regenerated with a small bounded retry budget. Finally, approved ones are stored in a reusable question bank for subsequent assessments.

\subsection{Fixed Context Grading}
\label{subsec:method_packet}

For each answer grading instance, we package the inputs as a fixed tuple:
{\setlength{\abovedisplayskip}{3pt}
 \setlength{\belowdisplayskip}{3pt}
 \setlength{\abovedisplayshortskip}{3pt}
 \setlength{\belowdisplayshortskip}{3pt}
\begin{equation}
u_i = (q_i, r_i, a_i),
\end{equation}}
where $q_i$ is the question, $r_i$ is a reference answer, and $a_i$ is the student answer (one of the four variants: \textsc{Good/Partial/Plausible/Bad} in our dataset). This representation standardises grading across subjects and languages and supports consistent evaluation across both proprietary and open-weight graders.

\paragraph{Bilingual Grading and Feedback Module.}
\label{subsec:method_grader}

Given a fixed tuple $u_i = (q_i, r_i, a_i)$, the grader ($Gr$) produces a numeric score and feedback:
{\setlength{\abovedisplayskip}{3pt}
 \setlength{\belowdisplayskip}{3pt}
 \setlength{\abovedisplayshortskip}{3pt}
 \setlength{\belowdisplayshortskip}{3pt}
 \begin{equation}
(\hat{s_i},\hat{f_i}) = Gr(q_i, r_i, a_i), \qquad \hat{s_i}\in[0,10].
\label{eq:grader}
\end{equation}}
Here, $\hat{s}_i$ is the graded score and $\hat{f}_i$ is the generated feedback. To ensure language consistency, we generate feedback in the same language as the question. The system returns outputs in a structured format (score, feedback text, and assessment summary) to support progress reporting and overall performance analysis; validated questions are stored in a reusable question bank.

Together, these components form an end-to-end pipeline for bilingual assessment, from heterogeneous inputs to question sets and graded written answers with appropriate feedback.

\section{Dataset and Fine-tuning}
\subsection{Preparing the Dataset}
\label{subsec:dataset}
\begin{table*}[t]
  \centering
  \includegraphics[width=\textwidth]{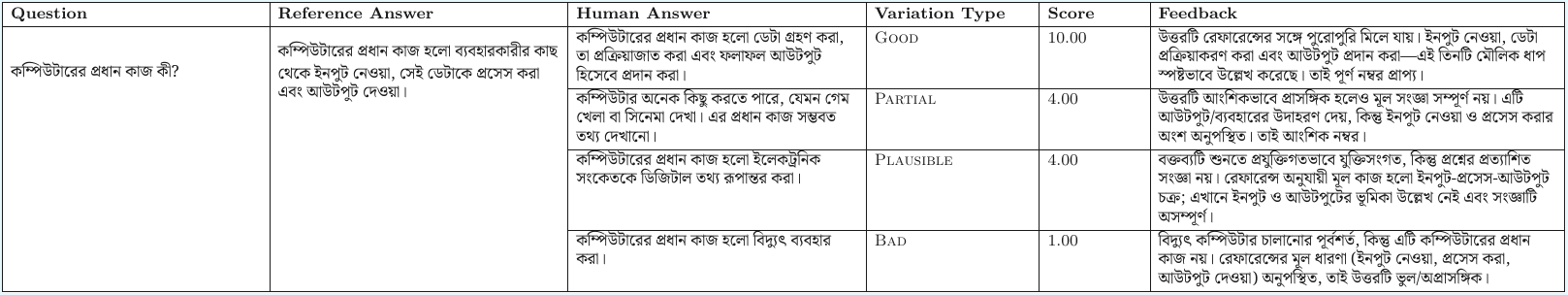}
  \caption{Example dataset record for one question, showing four variants of student answers with scores and feedback.}
  \label{tab:dataset_sample}
\end{table*}
\paragraph{Why Synthetic Data?} We build a synthetic dataset for controlled grading beyond binary grading by generating multiple student-answer variants per question–reference pair and assigning each variant a score and brief feedback. We use LLM-based synthesis because Bangla curricular materials (Grades 6--12) are scarce in standardized digital form; rubric-aligned annotation with explanations is costly. Via synthesis, for each $(q,r)$, we have generated a synthetic student answer:
{\setlength{\abovedisplayskip}{3pt}
 \setlength{\belowdisplayskip}{3pt}
 \setlength{\abovedisplayshortskip}{3pt}
 \setlength{\belowdisplayshortskip}{3pt}
 \begin{equation}
\begin{aligned}
\tilde{a}^{(v)} &= Gn(q,r,v),\\
&\forall\, v \in \{\textsc{Good, Partial, Plausible, Bad}\}.
\end{aligned}
\end{equation}} where $Gn$ is the generator. Finally we grade them with an external LLM grader $Gr_{ext}$:
{\setlength{\abovedisplayskip}{3pt}
 \setlength{\belowdisplayskip}{3pt}
 \setlength{\abovedisplayshortskip}{3pt}
 \setlength{\belowdisplayshortskip}{3pt}
 \begin{equation}
(\tilde{s}^{(v)}, \tilde{f}^{(v)}) = Gr_{ext}(q,r,\tilde{a}^{(v)}).
\end{equation} 
}
Here, \textsc{Good} answers are correct and complete, \textsc{Partial} answers are incomplete but partially correct; \textsc{Plausible} answers are lexically similar to the reference yet conceptually wrong, and \textsc{Bad} answers are largely incorrect or irrelevant.

\paragraph{Dataset Construction (Models, Prompting, and Decoding).}
To construct the dataset, we use a two-stage setup that separates question-answer synthesis from score assignment to reduce self-consistency bias: given Bangla resources, \textsc{Gemini 2.0 Flash} ($Gn$) generates questions and variant-diverse student-answers, while \textsc{Gemini 2.5 Flash} ($Gr_{ext}$) assigns scores and feedback. We instruct the model to be concise and apply a small, bounded retry policy, selecting the shorter valid feedback when alternatives are available, keeping all feedback under 256 tokens. This decoupling lets one model generate diverse student answers, while a separate model scores them using a fixed rubric, making the resulting labels more consistent. Each instance is stored as a structured record (question, reference, student answer, score, and feedback) to support deterministic parsing. Table~\ref{tab:dataset_sample} shows an instance of the dataset, and a prompt template is provided in the Appendix~\ref{app:prompts}.

\paragraph{Bilingual Coverage and Feedback Language.}
The dataset is bilingual, comprising 68.7\% Bangla and 31.3\% English question-answer pairs. We require the user's answer and the grader's feedback to be written in the same language as the question, which improves usability in classroom settings and avoids cross-lingual drift (e.g., Bangla inputs producing English feedback).

\paragraph{Dataset Integrity, Scale, and Splits.}
We store each question–reference pair as a structured record containing all four answer variants, their associated scores and feedback. While generating each instance, we apply lightweight validation checks to reduce synthesis noise and regenerate only failed components under a capped retry budget. The final dataset contains 122{,}710 questions, each with four graded answers (resulting in 490{,}391 instances, slightly fewer than expected due to failed generations), in total $\sim$112M tokens ($\sim$638.9\,MB on disk). Most instances fit within a 1024-token budget. We keep a fixed subset of $N{=}1000$ question-reference pair records for testing and split the remaining data into training/validation (90/10). We report model-to-model comparisons on the test subset, and from this same subset, a teacher-graded subset of $N{=}259$ records is used to measure alignment with human judgement.

\subsection{Fine-tuning in Low Resource Setting}
\label{sec:models}

\paragraph{Choosing the LLM Grader.}
Our system needs reliable \emph{reasoning}: the grader must compare a student's answer against the reference answer, assign partial credit when appropriate, and generate feedback that explains the judgement from the given resource rather than relying on surface word overlap. Hence, we adopt \textsc{Qwen3-8B} as our grader backbone because of its strong reasoning behaviour. Finally, we pick the 8B parameters scale as a practical middle ground: large enough for bilingual, rubric-conditioned semantic judgement, but small enough to support iterative fine-tuning and repeated evaluation runs during system development.

\paragraph{Fine-tuning Our Grader.}
To adapt the grader to our rubric and bilingual setting, we fine-tune our model using QLoRA \citep{dettmers2023qlora}: the backbone is loaded in 4-bit and kept frozen, while only low-rank LoRA adapter parameters are updated. This substantially reduces VRAM requirements, making fine-tuning feasible on commodity GPUs while retaining the backbone's general competence. We report full fine-tuning hyperparameters in Appendix~\ref{app:fine_infer}.
 
\section{Experiments}
\begin{table*}[t]
\centering
\small
\setlength{\tabcolsep}{3pt}
\renewcommand{\arraystretch}{0.95}
\resizebox{\textwidth}{!}{%
\begin{tabular}{lcccccccc}
    \hline
    \textbf{Model} &
    \textbf{Human $\rho$ $\uparrow$} &
    \textbf{Human MAE $\downarrow$} &
    \textbf{RoRa $\uparrow$} &
    \textbf{ReCEval $\uparrow$} &
    \textbf{ROSCOE $\uparrow$} &
    \textbf{IBE $\uparrow$} &
    \textbf{Reasoning Mark $\uparrow$} &
    \textbf{Token Length $\downarrow$} \\
    \hline
    Qwen3-8B (base)        & 0.904 & 1.174 & 0.790 & 0.458 & 0.727 & \textbf{0.721} & 6.74 & 93  \\
    \textbf{Qwen3-8B Fine-tuned (Ours)}  & \textbf{0.936} & \textbf{0.725} & \textbf{0.819} & \textbf{0.471} & 0.715 & 0.714 & \textbf{6.80} & \textbf{89}  \\
    GPT-5.2            & 0.902 & 1.112 & 0.609 & 0.413 & 0.734 & 0.715 & 6.18 & 147 \\
    Gemini 3 Flash     & 0.910 & 1.120 & 0.776 & 0.439 & \textbf{0.752} & 0.712 & 6.70 & 126 \\
    Claude Sonnet 4            & 0.919 & 0.903 & 0.653 & 0.428 & 0.703 & 0.688 & 6.18 & 164 \\
    Sonar Reasoning    & 0.892 & 0.992 & 0.667 & 0.464 & 0.684 & 0.638 & 6.13 & 196 \\
    LLaMA 3.2             & 0.778 & 1.752 & 0.674 & 0.432 & 0.692 & 0.715 & 6.28 & 143 \\
    \hline
  \end{tabular}
  }
  \caption{\label{tab:main_results}
  \textbf{Overall performance and reasoning-metrics comparison across graders:} Human agreement with \emph{(Spearman $\rho$ and MAE)}, reasoning quality of feedback with \textsc{RoRa}, \textsc{ReCEval}, \textsc{ROSCOE}, \textsc{IBE} and \emph{Reasoning Mark} (average of the previous four metrics scaled by x10), length of tokens with \emph{Token Length}. \textbf{Bold} marks the best score per column (ties allowed). ($\uparrow$ higher is better, $\downarrow$ lower is better.)}
\end{table*}
\subsection{Experimental Setup}
\label{subsec:exp_setting}
We create a synthetic bilingual Bangla–English dataset (638.9 MB on disk) and use it for task-specific QLoRA fine-tuning of an open-weight LLM (Qwen3-8B) as our grader. Fine-tuning is performed on $2\times$RTX~3090 GPUs, and we serve the resulting grader on a single RTX~3090 (24GB) for inference. We evaluate the fine-tuned grader against other open-weight and proprietary baselines on a primary held-out set of $N{=}1000$ instances under a fixed input/output interface for fair comparison. Finally, to quantify the remaining gap to human judgment and contextualize performance gains, we conduct a dedicated human study to measure alignment with a separate teacher-graded subset of $N{=}259$ instances.



\subsection{Evaluation}
\label{sec:evaluation}

We evaluate along two axes: \emph{(i) score fidelity with the human study}: alignment with teacher grades and \emph{(ii) feedback quality}: whether the feedback provides evidence-based justification instead of paraphrasing the reference answer.

\paragraph{Compared Baseline Models.}
To contextualize our grader against widely deployed systems, we compare against several proprietary LLM graders: \textsc{Gemini 3 Flash}, \textsc{GPT-5.2}, \textsc{Claude Sonnet 4}, \textsc{LLaMA 3.2}, and \textsc{Sonar Reasoning} (via the Perplexity API). These serve as \emph{API-only} baselines with no access to model weights or training details and represent strong graders used in practice; all are evaluated under the same fixed context and a shared $[0,10]$ scoring scale.

\paragraph{Human Evaluation.}
We analyze model-graded scores against teacher grades on a held-out set of $N{=}259$ responses, stratified by subject and language (Bangla/English). We recruited 20 school teachers; each response is scored once by a single teacher on a $[0,10]$ scale. Then, we compare model scores to teacher scores using Spearman correlation \citep{spearman1904proof} ($\rho$ : measures whether models preserve the same ordering of students' answers, even if score levels differ slightly), mean absolute error (MAE : average absolute point deviation, easy to interpret in points and less sensitive to outliers than MSE) \cite{willmott2005advantages}.

\paragraph{Automatic Feedback Metrics.}
Beyond score agreement, we evaluate the quality of each model’s feedback using four complementary measures: \textsc{ReCEval} \citep{prasad-etal-2023-receval} (logical consistency and informativeness), \textsc{ROSCOE} \citep{golovneva2023roscoe} (semantic/faithfulness cues), \textsc{IBE-Eval} \citep{dalal2024ibe} (overall explanation quality), and \textsc{RoRa} \citep{rora2024} (leakage resistance). All these metrics are language agnostic, and we compute them on the same evaluation set and report per-model aggregates. \textsc{RoRa} is our primary leakage diagnostic: it measures to what extent the feedback provides an independent justification for the score, rather than paraphrasing the reference answer or directly encoding the score. To avoid self-evaluation bias, these metrics are computed with independent evaluator backbones (an MNLI-trained NLI model \citep{williams-etal-2018-broad,lewis-etal-2020-bart} and a multilingual sentence-embedding model \citep{reimers-gurevych-2019-sentence, song-etal-2020-mpnet}).  Finally, we report the unweighted mean of these metrics (\emph{Reasoning Mark}) as a single aggregate indicator.

\paragraph{Evaluating Token Length.}
To control for verbosity, we measure feedback length in output tokens. Token length directly reflects generation cost and latency (i.e., the number of decoding steps). We report the per-model mean of token length to ensure improvements are not mainly driven by only longer outputs.

\subsection{Results and Analysis}
\label{results}

\begin{figure}[!htbp]
    \centering
    \includegraphics[width=\linewidth]{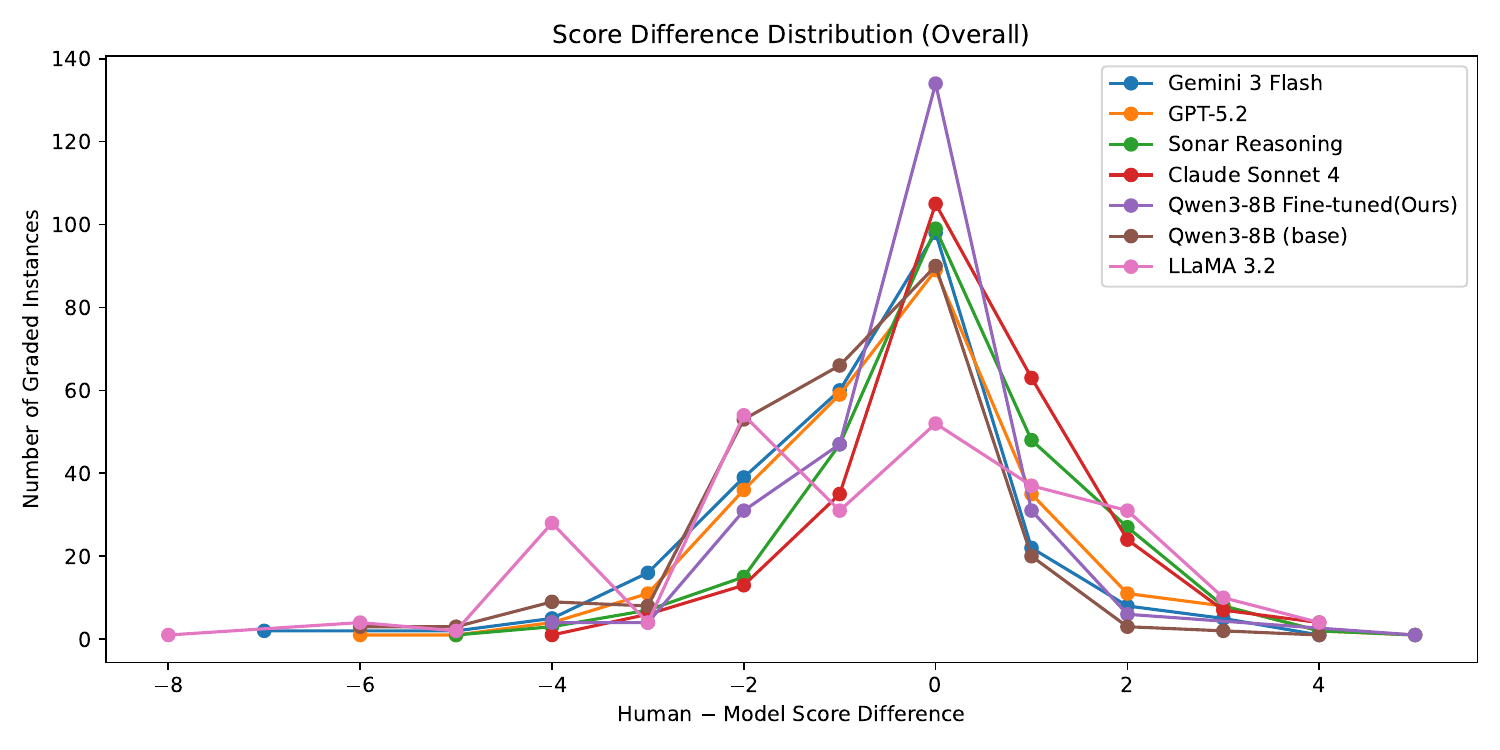}
    \caption{\textbf{Teacher--model score difference distribution across graders.} The x-axis shows the score difference (teacher$-$model) and the y-axis shows the number of graded instances. Values near 0 indicate closer agreement with teacher scoring, while heavier tails indicate more frequent large deviations.} 
    \label{fig:score_diff}
\end{figure}

Table~\ref{tab:main_results} shows that task-specific QLoRA fine-tuning on our synthetic bilingual dataset improves scoring alignment with teacher grades: \textbf{Fine-tuned Qwen3-8B leads the compared graders on human-alignment metrics}. We then examine the error distribution (Figure~\ref{fig:score_diff}), feedback quality, and efficiency to explain where these gains come from.

\paragraph{Agreement with Human Teachers.}
\label{subsec:human_agreement}

We compare graders using two alignment signals: rank agreement (Spearman $\rho$) and absolute error (MAE) against teacher scores. As shown in Table~\ref{tab:main_results}, fine-tuned Qwen3-8B achieves the strongest human alignment ($\rho{=}0.936$, MAE$=0.725$) among all models and improves over Qwen3-8B (base) ($\rho{=}0.904$, MAE$=1.174$). This indicates better agreement with teachers on which student answers deserve higher scores, while also reducing absolute deviation from teacher-assigned scores on the same [0,10] scale. To analyze where this gain comes from, Figure~\ref{fig:score_diff} plots the distribution of score differences (teacher$-$model). \textbf{Fine-tuned Qwen3-8B shows the tightest concentration around zero and the lightest tails, suggesting fewer extreme over/under-scoring errors among all compared graders}. Overall, task-specific fine-tuning improves both ranking consistency and point-level accuracy while maintaining concise generation, demonstrating that alignment gains are achieved without increasing output verbosity.

\paragraph{Reasoning Analysis.}
\label{subsec:finetuning_rora}

We compare feedback quality using leakage-aware and reasoning metrics (\textsc{RoRa}, \textsc{ReCEval}, \textsc{ROSCOE}, \textsc{IBE}), together with output length (tokens) (Table~\ref{tab:main_results}). As shown in Table~\ref{tab:main_results}, fine-tuned Qwen3-8B improves over Qwen3-8B (base) on leakage resistance (\textsc{RoRa}: 0.790$\rightarrow$0.819) and reasoning faithfulness (\textsc{ReCEval}: 0.458$\rightarrow$0.471), while staying concise (93$\rightarrow$89 tokens). These gains indicate that fine-tuning produces more answer-specific, non-leaked feedback without increasing verbosity. Across all graders, fine-tuned Qwen3-8B is \textbf{best} on \textsc{RoRa} (0.819) and \textsc{ReCEval} (0.471), and remains close to the top on \textsc{ROSCOE} (0.715) and \textsc{IBE} (0.714). Overall, it ranks best in \textbf{6 of 8} reported metrics in Table~\ref{tab:main_results}, showing that quality gains are broad rather than coming from a single metric.

\paragraph{Token Efficiency.}
\label{subsec:efficiency}
We compare graders by average output length in tokens. As shown in (Table~\ref{tab:main_results}), \textbf{fine-tuned Qwen3-8B is the most token-efficient grader (89 tokens)}, slightly below Qwen3-8B (base) (93) and far below proprietary models such as GPT-5.2 (147) and Gemini 3 Flash (126). This efficiency reduces per-response computation and latency, and it comes \emph{without} sacrificing alignment or leakage-resistant feedback quality.

\section{Demonstration}
\label{sec:demo}We deploy our system as a web-based assessment interface; its main components are illustrated as follows. Once users are taken to the exam interface, they can upload a typed or handwritten text image and submit a grading request (Figure~\ref{fig:demo_htr}). The system returns a numeric score together with a short, context-grounded feedback (Figure ~\ref{fig:demo_grading}). To support practical classroom use, the demo also exposes a summary of the users' total marks and overall performance (Figure ~\ref{fig:demo_summary}).

{\setlength{\floatsep}{4pt}
 \setlength{\textfloatsep}{6pt}
 \begin{figure}[t]
  \centering
  \includegraphics[width=\linewidth]{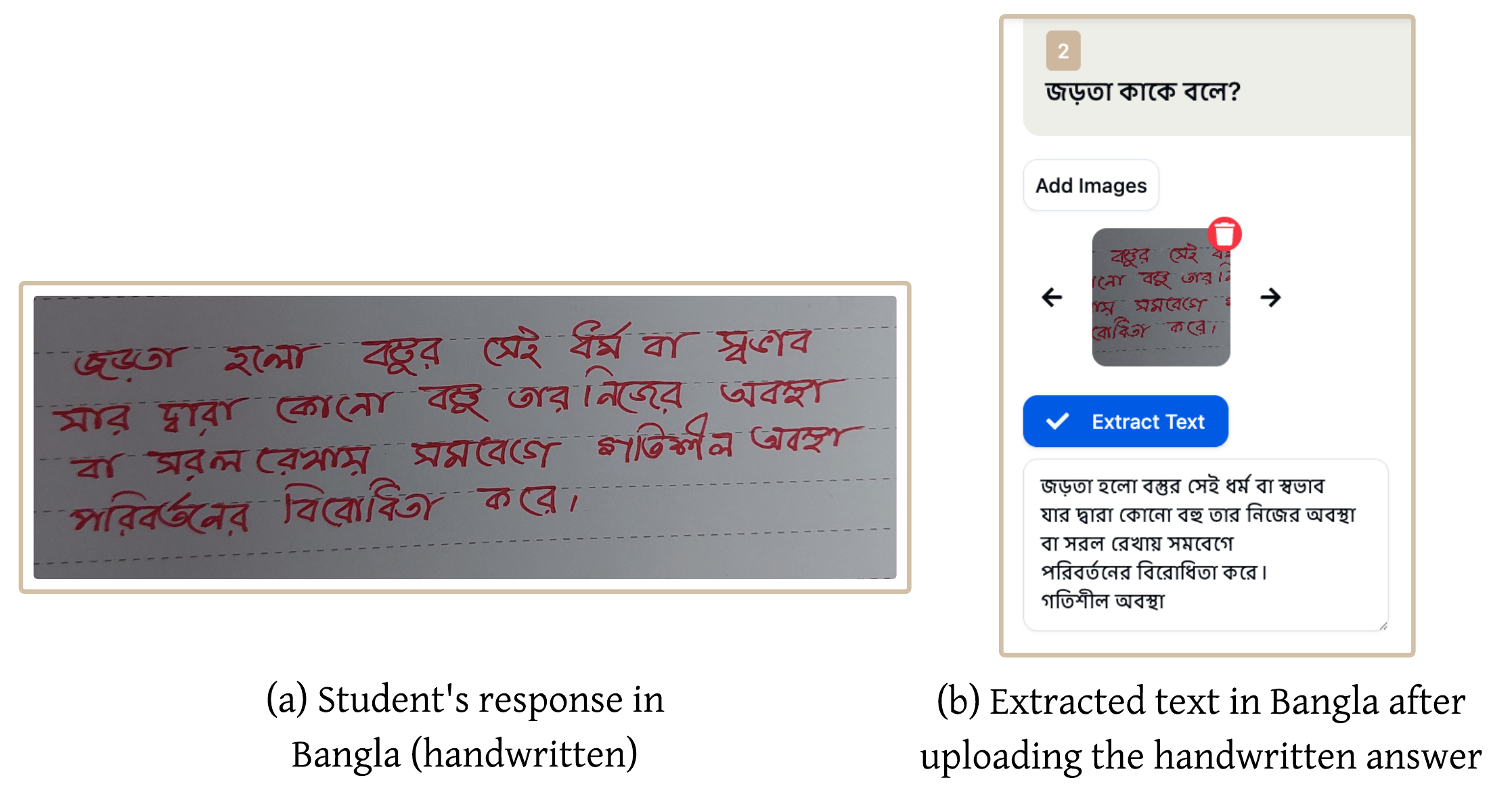}
  \caption{\textbf{Answer input:} (a) A student can either type or upload a handwritten response image; (b) the integrated Bangla HTR module extracts text for downstream grading.}
  \label{fig:demo_htr}
\end{figure}

\begin{figure}[t]
  \centering
  \includegraphics[width=\linewidth]{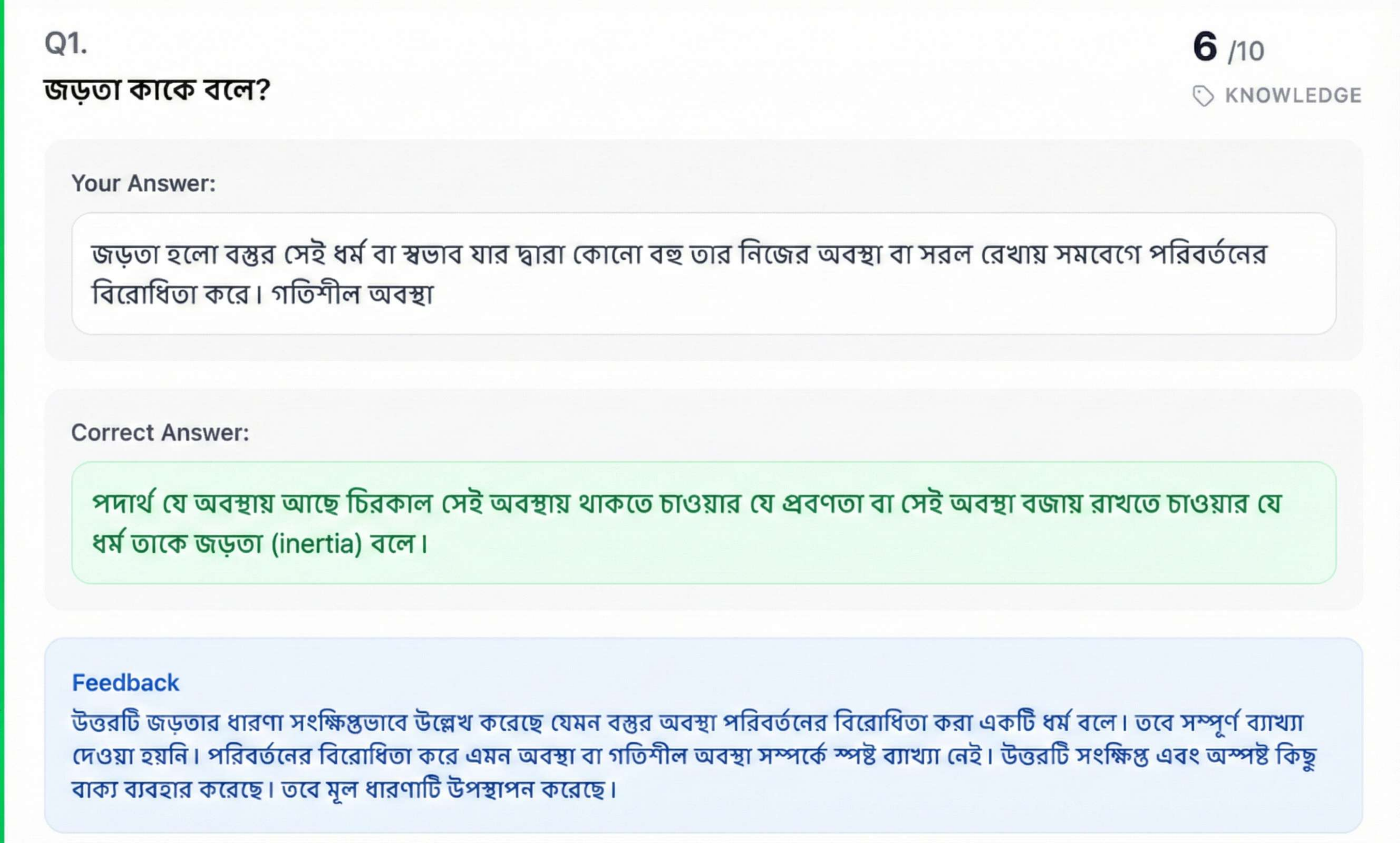}
  \caption{\textbf{Grading and feedback view:} The interface shows the student's answer, the reference answer, and the model’s predicted score with brief feedback defending the given score.}
  \label{fig:demo_grading}
\end{figure}

\begin{figure}[t]
  \centering
  \includegraphics[width=\linewidth]{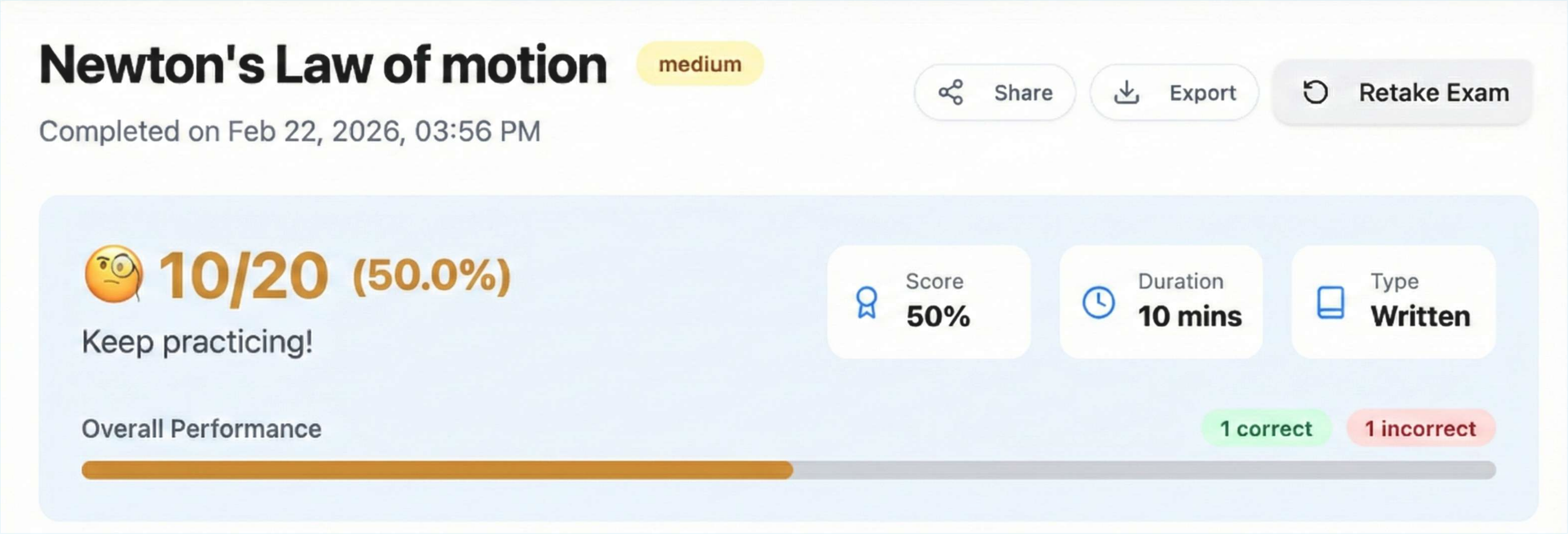}
  \caption{\textbf{Exam-level summary:} Aggregated performance over an exam session.}
  \label{fig:demo_summary}
\end{figure}
}

\section{Conclusions}
We present a bilingual (Bangla and English) written-answer evaluation system that produces human-aligned scores together with reliable, context-grounded feedback. To move beyond surface-overlap grading, we introduce a controlled four-variant dataset and fine-tune an open-weight language model to support semantically grounded evaluation. Across experiments, our system improves agreement with teacher-assigned scores and yields more faithful explanations than strong proprietary graders, as confirmed by a dedicated human evaluation study. Overall, we show that an accurate, trustworthy automated assessment for low-resource classrooms is achievable with open models when grading is explicitly grounded in the question and reference answer and paired with bounded, context-aware feedback.


\section*{Limitations and Future Work}
Our system has three main limitations. First, part of our dataset relies on LLM-generated answer variants, scores, and feedback, which may introduce synthetic artifacts and reflect the broader scarcity of large-scale teacher-annotated Bangla exam data with expert rationales. Second, the dataset is not explicitly stratified by curriculum objectives or cognitive levels (e.g., Bloom’s taxonomy), limiting fine-grained control over difficulty and skill coverage. Third, the current system does not support multimodal exam items that require reasoning over diagrams, tables, or other visual evidence. Future work can focus on scaling teacher-annotated Bangla data collection, incorporating curriculum- and Bloom-aware controls for grading and feedback, and extending the framework to visually grounded assessment with systematic evaluation protocols. 

\section*{Ethical Considerations}
Our system is intended to assist educators with question generation and grading, not to replace teacher judgment. As with any automated grader, it can produce incorrect scores or feedback, particularly for ambiguous answers, domain shifts, or degraded OCR/HTR outputs; we therefore recommend human oversight for any high-stakes use.

Student responses and exam artifacts may contain sensitive personal information. Deployments should follow strict data-handling practices, including data minimization, limited retention, access control, and secure storage and transmission. We do not support uses of the system for surveillance, profiling, or punitive decision-making.

Automated grading can also amplify biases (e.g., against dialectal, non-standard, or stylistically different writing). Future work should include targeted evaluation across demographic and linguistic variations, and provide practical auditing and override mechanisms so teachers can inspect, correct, and contest system outputs.

\section*{Acknowledgements}
We acknowledge Department of Computer Science and Engineering, University of Dhaka, Bangladesh, for computational support. At the same time, we are grateful to the participating teachers for providing graded responses used in the evaluation. 

\bibliography{custom}

\appendix
\section{Appendix}

\subsection{Prompt Templates and Output Schemas}
\label{app:prompts}

This appendix documents the prompt templates used in our pipeline.
We enforce \textbf{strict JSON-only outputs} to enable deterministic parsing and
to reduce annotation noise in large-scale generation and evaluation.

\paragraph{Synthetic Answer-Variant Generation Prompt.}
We generate four controlled student-answer variants (\textsc{good/partial/plausible/bad})
conditioned on the question and reference answer. The model must return a JSON array
containing exactly four objects, each with a \texttt{type} label and a \texttt{user\_answer} string.

\paragraph{Prompt:}\mbox{}
\begin{lstlisting}
You are an assistant generating possible user answers to a question.

Question:
{question}

Reference Answer (gold standard):
{reference_answer}

Generate FOUR distinct 'user_answer' variations in JSON format:
1. A good answer (nearly perfect)
2. A partially correct answer (captures some ideas but misses key parts)
3. An incorrect but plausible answer (sounds right but is wrong)
4. A bad answer (completely off-topic or nonsense)

Return in JSON array format like:
[
  { "type": "good", "user_answer": "..." },
  { "type": "partial", "user_answer": "..." },
  { "type": "plausible", "user_answer": "..." },
  { "type": "bad", "user_answer": "..." }
]
\end{lstlisting}

\paragraph{Required JSON schema:}\mbox{}
\begin{lstlisting}
[
  {"type": "good"|"partial"|"plausible"|"bad", "user_answer": string},
  ...
]
\end{lstlisting}

\paragraph{Synthetic Evaluation (Score + Feedback) Prompt:}
Each generated answer is then scored against the same evidence packet
(\texttt{question}, \texttt{reference\_answer}, \texttt{user\_answer}).
We constrain the output to a JSON object with a numeric \texttt{score} and a free-text
\texttt{feedback}. We also require the feedback language to match the question language.

\paragraph{Prompt:}\mbox{}
\begin{lstlisting}
You are an expert grader. Assess a user's answer against a Reference Answer.

Provide:
1. A numerical score (0-10)
2. Detailed reasoning for your score.
Make sure the reasoning language is in the provided question's language
and output nothing except the reasoning.

Return strictly in JSON format:
{
  "score": <0-10>,
  "reasoning": "<reasoning>"
}

Question:
{question}

Reference Answer:
{reference_answer}

User Answer:
{user_answer}
\end{lstlisting}

\paragraph{Required JSON schema:}\mbox{}
\begin{lstlisting}
{"score": integer (0..10), "reasoning": string}
\end{lstlisting}

\paragraph{Final Grading Prompt Used by the Fine-Tuned Grader.}
For training the open-weight grader and for inference-time grading, we use the same evidence-packet input fields. The prompt is identical to our \emph{Synthetic Evaluation (Score + Feedback) Prompt} and is reused both during QLoRA fine-tuning and at inference time.

\paragraph{Training target format.}
During fine-tuning, each training instance packs the above prompt as input and uses a
JSON string target \texttt{\{"score": int, "reasoning": string\}} for supervision.

\paragraph{Parsing and validation.}
We reject or retry generations that are not valid JSON or that omit required keys.
This enables robust, large-scale training data construction and consistent grader outputs.

\subsection{Output Validation and Bounded Retry Policy}
\label{app:validation_retry}
We apply lightweight output validation to every grader response to ensure a stable JSON interface and to reduce malformed or overly verbose generations. Concretely, we require a single parseable JSON object with exactly \texttt{score} and \texttt{feedback}; \texttt{score} must be numeric in $[0,10]$, \texttt{feedback} must be a non-empty string written in the same language as the question (Bangla/English), and the reasoning must stay within a fixed token budget. If validation fails, we re-prompt with an explicit correction instruction under a bounded retry policy ($K{=}2$), and when multiple valid outputs are available we select the shortest one by token count to minimize latency and decoding cost.

\subsection{Additional dataset statistics}
\label{app:dataset_stats}
Average token lengths are 23.51 (question), 52.43 (reference), 35.46 (student answer), and 71.93 (feedback), with an average combined sequence length of 228.32 tokens and a maximum length of 1448 tokens. Most instances fall within a 1024-token budget (490{,}384 instances), with only a small remainder exceeding it.

\subsection{Fine-tuning and Inference Settings}
\label{app:fine_infer}

\noindent\textbf{Fine-tuning.}
We fine-tune \texttt{Qwen/Qwen3-8B} using QLoRA-style PEFT: the backbone is loaded in 4-bit (\texttt{load\_in\_4bit=True}) with NF4 quantization, double quantization, and FP16 compute, and kept frozen while training LoRA adapters (causal LM) with $r{=}16$, $\alpha{=}32$, dropout 0.05. We use AdamW with learning rate $2\times10^{-5}$ for 1 epoch, max sequence length 1024, per-GPU batch size 1 with gradient accumulation 8 (effective batch 16 on $2\times$RTX~3090). We enable mixed precision (BF16 when available, otherwise FP16), disable KV caching (\texttt{use\_cache=False}), and enable gradient checkpointing to reduce memory. Data are split 90\%/10\% into train/validation with a fixed seed (42), and we select checkpoints by minimum validation loss (saving the best; \texttt{save\_last=True}).

\noindent\textbf{Inference.}
Local grader decoding uses \texttt{max\_new\_tokens=1024} and \texttt{temperature=0.2}. For API baselines, we use \texttt{temperature=0.2} and \texttt{max\_tokens=800} (where applicable). We apply bounded retries (up to 3) for malformed or non-JSON outputs and keep the shortest valid completion when multiple candidates are available.

\subsection{Teacher evaluation protocol and metrics}
\label{app:teacher_eval}
We evaluate score alignment against teacher grading on a held-out set of $N{=}259$ responses, stratified by subject and language (Bangla/English). We recruited 20 school teachers with subject-matter expertise; each response is scored once by a single teacher on a $[0,10]$ scale. For a model $m$ producing predicted scores $\hat{s}^{(m)}_i$ and teacher scores $s_i$, we report (i) rank agreement via Spearman correlation $\rho$ and (ii) absolute deviation via mean absolute error (MAE):
\begin{equation}
\rho^{(m)}=\mathrm{Spearman}\big(\{\hat{s}^{(m)}_i\}_{i=1}^{N}, \{s_i\}_{i=1}^{N}\big),
\end{equation}
\begin{equation}
\mathrm{MAE}^{(m)}=\frac{1}{N}\sum_{i=1}^{N}\left|\hat{s}^{(m)}_i-s_i\right|.
\end{equation}

\subsection{Feedback length statistics}
\label{app:feedback_length}
We measure feedback length in \emph{output tokens}. For each model $m$, let $\ell^{(m)}_i$ be the feedback token count for instance $i$. In the main paper, we report the \textbf{per-model mean} token length,
\begin{equation}
\mu^{(m)}=\frac{1}{N}\sum_{i=1}^{N}\ell^{(m)}_i,
\end{equation}
and we also compute the standard deviation for completeness,
\begin{equation}
\sigma^{(m)}=\sqrt{\frac{1}{N-1}\sum_{i=1}^{N}\bigl(\ell^{(m)}_i-\mu^{(m)}\bigr)^2}.
\end{equation}

\end{document}